\documentclass{article}
\usepackage{spconf,amsmath,graphicx,multirow}
\usepackage[table,xcdraw]{xcolor}
\usepackage{amssymb}

\title{DEFRAG: Deep Euclidean
Feature Representations through Adaptation on the Grassmann Manifold}

\name{Breton Minnehan and Andreas Savakis}
\address{Rochester Institute of Technology\\
	Rochester, New York 14623, USA}

\begin{document}
%
\maketitle
\begin{abstract}
We propose a novel technique for training deep networks with the objective of obtaining feature representations that exist in a Euclidean space and exhibit strong clustering behavior. Our desired features representations have three traits: they can be compared using a standard Euclidian distance metric, samples from the same class are tightly clustered, and  samples from different classes are well separated. However, most deep networks do not enforce such feature representations.  The DEFRAG training technique consists of two steps: first good feature clustering behavior is encouraged though an auxiliary loss function based on the Silhouette clustering metric. Then the feature space is retracted onto a Grassmann manifold to ensure that the $L_2$ Norm forms a similarity metric. The DEFRAG technique achieves state of the art results on standard classification datasets using a relatively small network architecture with significantly fewer parameters than many standard networks. 
\end{abstract}
\begin{keywords}
Deep learning, Clustering, Grassmann Optimization
\end{keywords}
\section{INTRODUCTION}
\label{sec:intro}

Deep Convolutional Neural Networks (CNNs) and their variants have emerged as the architecture of choice for computer vision. Deep networks have achieved state-of-the-art results in object class recognition \cite{ksh12}, \cite{sz15}, \cite{hzrs16}, face recognition \cite{skp15}, semantic segmentation \cite{lsd15}, pose estimation \cite{wrks}, and visual tracking \cite{nh15} among other applications.  While the initial focus has been on making CNNs deeper in order to achieve higher performance, recent work has been exploring leaner networks, such as DenseNet \cite{hlw16} as an alternative to ResNet \cite{hzrs16}, that are more efficient yet perform just as well if not better than their larger counterparts.  

\begin{figure}[t]
\centerline{\includegraphics[width=60mm]{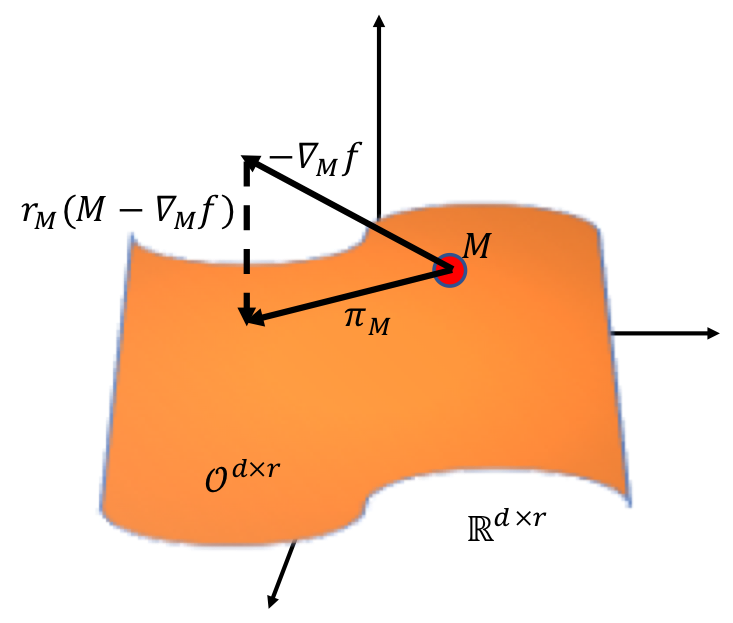}}
\caption{Visualization of DEFRAG process of  optimization along the gradient direction and then retraction on the Grassmann manifold.}
\label{fig:grassRetract}
\end{figure}

In its simplest form a CNN consists of a feature extraction convolutional network followed a linear classifier, the head of the network. One benefit of CNNs is that they are trained in an end to end manner, thus the maximum benefit can be extracted from each stage. However, the features learned by CNNs can be further improved for robustness. A robust feature representation is one that minimizes differences between samples of the same class and maximizes differences between samples from different classes. In this work we present a method referred to as DEFRAG, inspired by Linear Discriminate Analysis (LDA) \cite{fisher1936use}.  Our approach shapes the feature representation through a novel auxiliary loss function, (Section \ref{ssec:euAuxLoss}), and a orthogonalization step that involves retraction on a Grassmann manifold (Section \ref{ssec:grMfnRtr}) illustrated in Fig. \ref{fig:grassRetract}.

An auxiliary loss, $L_{aux}$, is secondary metric that is added to the loss from the main training objective, $L_{class}$, for the optimizer to minimize, as 
\begin{equation}
L=L_{Class}+\lambda_{aux} L_{aux}
\end{equation}
Where $\lambda_{aux}$ is a mixture parameter used to balance the impact of the auxiliary loss. 

In this work the main loss objective is the traditional categorical cross-entropy loss which learns to classify the samples, and a proposed new auxiliary loss function, the Silhouette Loss.  Figure \ref{fig:full2DLoss} illustrates the clustering characteristics of DEFRAG compared to other methods.

The contributions of this paper are the following: (a) We introduce a new auxiliary loss functions based on the Silhouette clustering metric which encourages tight intra-class clustering and inter-class separation. (b) We propose an orthogonalization step which retracts the optimized feature projection matrix back on the Grassmann manifold. (c) We demonstrate how the DEFRAG method improves  performance so that a small network matches or exceeds the performance of much bigger networks and achieves state of the art results on standard datasets. 

In the remainder of this paper, Section 2 provides background on auxiliary loss functions and other regularization methods used to train deep neural networks. Section 3 presents our proposed DEFRAG method and its theoretical justifications. Section 4 outlines our results and Section 5 offers concluding remarks.

\section{BACKGROUND}
\label{sec:background}

There are many variants of auxiliary loss functions used when training deep networks to encourage different behaviors. One of the first auxiliary losses proposed was feature regularization. The goal of regularizing the feature activations is to keep the values in the feature representation small or sparse by using $L_2$ or $L_1$ norms respectively. The underlying assumption is that small-valued or sparse feature representations generally reduce over-fitting. $L_2$ regularization encourages  activations with small magnitudes
\begin{equation}
L_2=\lVert G(\mathbf{x}\mathbf{W}+\mathbf{b})\rVert _2^2
\end{equation}
Where $G(\cdot)$ is an activation function, $\mathbf{x}$ is the layer's input vector, $\mathbf{W}$ is the weight matrix 
and $\mathbf{b}$ is the bias vector. 
$L_1$ regularization encourages sparsity in the activations
\begin{equation}
L_1=\lVert G(\mathbf{x}\mathbf{W}+\mathbf{b})\lVert _1
\end{equation}

The use of these regularization techniques has waned due to reliance on robust training methods such as Batch Normalization \cite{is15} and Dropout \cite{shkss14}. Batch Normalization reduces the covariance shift and potential for vanishing or exploding gradients by normalizing the activations from each layer to be zero-mean and unit-variance. Dropout on the other hand reduces the potential for co-adaptation of neurons by randomly setting a fraction of neurons to zero on each training iteration, thus forcing neurons to be more self-reliant. Though these techniques have improved training time and network accuracy, they overlook generalizable feature representations.

Recent work introduced an auxiliary function called the center loss \cite{wzlq16}, that increases the robustness of the feature representation by encouraging tightly grouped clusters. The center loss represents the average distance of each point $\mathbf{x_i}$, in feature space, to the mean $\mathbf{c_{y_i}}$ of the corresponding class $y_i$. 
\begin{equation}
L_C=\frac{1}{2m} \sum_{i=1}^m \lVert \mathbf{x_i}-\mathbf{c_{y_i}} \rVert_2^2 
\label{eq:LC}
\end{equation}
\noindent
where m is the number of samples in the mini-batch, $\mathbf{x_i}$ is the feature space representation for the $i^{th}$ sample and $\mathbf{c_{y_i}}$ is the center for the class $y_i$ of the $i^{th}$ sample.

The tight Euclidean clustering in feature space encouraged by the center loss is useful in situations where the feature representations are compared to estimate similarity between samples, such as is done with k-Nearest Neighbors (k-NN).
The work in \cite{wzlq16} focused on person re-identification, a problem that requires robust feature representations that can be compared using Euclidean distance metrics. 

More recently \cite{qs17} introduced contrastive center loss, a technique that encourages tight clustering and increases class separation. 
The contrastive center loss is given as:
\begin{equation}
L_{CtC}=\frac{1}{m} \sum_{i=1}^m {\frac{\lVert \mathbf{x_i}-\mathbf{c_{y_i}} \rVert _2^2}{\sum_{j=i,j\neq y_i}^k \lVert \mathbf{x_i}-\mathbf{c_j} \rVert _2^2+\delta}}
\label{eq:LCtC}
\end{equation}
where $\delta$ is a small value that insures the denominator is non-zero. 
Our work builds upon the center loss and contrastive center loss for better feature clustering and more robust performance.

Like many other methods in the field, both the \cite{wzlq16} and \cite{qs17} use the Euclidean, $L_2$, norm as a similarity metric between deep features vectors. This  use of the $L_2$ norm can be problematic when applied to arbitrary feature spaces. The Euclidean norm is intended to operate in $\mathbb{R}^N$, represented by an orthogonal basis. However, the $L_2$ norm is often applied to vector representations without an orthogonal basis. In this work we aim produce feature representations with a proper $L_2$ similarity metric. This behavior is critical for many applications that use feature similarity, such as k-NN and other graphical methods. We insure orthogonality by retracting the weight matrix of our feature representation layer to the Grassmann manifold, the set of orthogonal spaces in $\mathbb{R}^N$. 

\section{Proposed Methodology}
\label{sec:method}
The DEFRAG method consists of two components: an auxiliary loss component, and a retraction of the feature projection on the Grassmann manifold. The auxiliary loss is designed to encourage better feature clustering of samples based on their class labels, while the Grassmann manifold retraction ensures the features are in a space suitable for the $L_2$ norm similarity metric. A side benefit of our DEFFRAG training approach is that due to the robustness of the features generated, smaller networks can be used. The two components of DEFRAF are discussed next.

\subsection{Clustering Auxiliary Loss }
\label{ssec:euAuxLoss}
Robust feature representations are important for  classification,  as they  increase the classifier's ability to generalize 
across different datasets and operating conditions. We formulate the formation of robust feature representations as seeking a feature space that encourages tight clusters for samples in the same class and large separations between clusters from different classes, for some similarity metric. Training deep networks with only the classification loss does not inherently encourage feature clustering.

Our auxiliary loss function is the Silhouette loss $L_{sil}$ shown below:  
\begin{equation}
L_{Sil}=\frac{1}{m} \sum_{i=1}^m {\frac{\lVert \mathbf{x_i}-\mathbf{c_{y_i}} \rVert_2^2}{min_{j\neq y_i}⁡\lVert \mathbf{x_i}-\mathbf{c_j} \rVert_2^2 +\delta}} 
\label{eq:LSil}
\end{equation}

The Silhouette loss is inspired by the Silhouette score \cite{r87}, which is used to assess clustering performance.The Silhouette loss is different from the center loss and contrastive center loss in that is focuses on separating classes that are close to each other, instead of maximizing the overall class separation. This alteration is important because it forces the network to focus on classes that are hard to separate and results in better classification performance.  
To reduce computation we use a running average method to update the class centers 
as suggested in \cite{wzlq16}.

\subsection{Grassmann Manifold Retraction}
\label{ssec:grMfnRtr}
The set of orthogonal spaces in $\mathbb{R}^N$ form a  Grassmann manifold. We therefore formulate our feature learning process as an optimization problem on the Grassmann manifold, our optimization process depicted in Fig. \ref{fig:grassRetract}. This is done by using a linear activation function for the last layer of the network, reducing it to a linear projection with projection matrix $M$. The update step of the Grassmann optimization is done using traditional stochastic gradient descent. The projection matrix is updated each iteration of training based on the gradient of the loss function $\nabla_Mf$. However, because any non-zero update is likely to take the projection matrix out of the manifold, the updated matrix is retracted back on the Grassmann manifold as, $r_M(M-\nabla_Mf)$. This retraction is done through singular value decomposition \cite{fan1995some} as, 
\begin{equation}
M'=UV^T, \textnormal{ where } M=USV^T.
\end{equation}
This process enforces features that satisfy the Silhouette auxiliary loss criterion and reside in an orthonormal space.

\begin{figure}[tbh]
\centerline{\includegraphics[width=80mm]{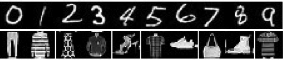}}
\caption{Example images from MNIST (top row) and Fashion MNIST (bottom row).}
\label{fig:examples}
\end{figure}
\section{Experimental Results}
\label{sec:Exp}

We performed a series of experiments by training a deep neural network using the proposed DEFRAG method and compared the results with the state of the art deep networks. The network architecture of our choice is small network with only two convolutional layers and three fully connected layers. The network parameters are given in Table 1. This network choice is intentionally simplistic so that the benefit of the DEFRAG method is made apparent. 

\begin{table}[tb]
\centering
\caption{Network Architecture}
\label{table:netArch}
\begin{tabular}{|l|r|r|}
\hline
Stage                                & Layer Type  & Size      \\ \hline
\multirow{3}{*}{Convolution Stage 1} & Conv        & 32 (5x5)  \\ \cline{2-3} 
                                     & Pooling     & 2x2       \\ \hline
\multirow{3}{*}{Convolution Stage 2} & Conv        & 256 (5x5)  \\ \cline{2-3} 
                                     & Pooling     & 2x2       \\ \hline
Fully Connected              		 & ReLU 	   & 256    \\ \hline
Feature Representation               & ReLU/Linear & 8    \\ \hline
Output                               & Softmax     & 10        \\ \hline
\end{tabular}
\end{table}

We used standard classification datasets in these experiments including: MNIST \cite{lbbh98} and Fashion MNIST \cite{xrv17}. Example images of the datasets used are shown in Fig. \ref{fig:examples}. In Section \ref{ssec:fashMNIST} we show  quantitative comparisons of the proposed DEFRAG method to standard deep networks on the Fashion MNIST dataset. Section \ref{ssec:qualMNIST} presents qualitative results of feature clustering with the MNIST \cite{lbbh98} dataset.

\subsection{Fashion MNIST Experiments}
\label{ssec:fashMNIST}
The Fashion MNIST dataset \cite{xrv17} was developed as a significantly more challenging alternative to the original MNIST dataset \cite{lbbh98}. Like the original MNIST dataset, Fashion MNIST dataset consists of 60,000 training samples and 10,000 test samples from objects in 10 different classes. The objects in the Fashion MNIST set are ten different articles of clothing: T-Shirt/Top, Trouser, Pullover, Dress, Coat, Sandals, Shirt, Sneaker, Bag, Ankle boots. This dataset is significantly harder than the original MNIST dataset. In crowed sourced experiments on a subset of 1000 examples humans were only able to achieve 83.5\% accuracy. The current state of the art performance on the dataset without augmentation is only 93.7\%. Examples form the Fashion MNIST dataset are shown on the bottom row of Figure \ref{fig:examples} and the experimental results are summarized in Table \ref{tbl:fashMNIST}. 

Our experiments with the Fashion MNIST dataset illustrate the impact on classification accuracy of both DEFRAG components, the Silhouette loss and orthogonalization step. 
We first trained the network using only the classification loss traditional ReLU activation function to get a baseline, Sparse (ReLU) in the table. We then considered a linear activation function for the feature representation layer, as well as Silhouette, Center and DEFRAG.
The results in Table \ref{tbl:fashMNIST} demonstrate that the DEFRAG method  outperforms the other methods with a relative reduction in error of 7\% compared the the original network. The results also demonstrate a significant improvement using just the Silhouette loss, however, DEFRAG shows improvement as a result of the orthonormal feature space. 

Our experiments demonstrate that state of the art results can be achieved on the Fashion-MNIST dataset with a simple network that benefits from a feature space clustering regularization technique. The accuracy achieved by our network is better than the results achieved with the much larger GoogLeNet \cite{sljsraevr14} and  VGG16 \cite{sz15} architectures.  In comparison with the other two architectures, our network has 24 and 1.4 times fewer parameters, respectively. These comparisons are based on the updated results reported in \cite{fash}.

\begin{table}[tb]
\centering
\caption{Fashion MNIST Results}
\label{tbl:fashMNIST}
\begin{tabular}{|p{2cm}|p{2.4cm}|p{1.1cm}|p{1.3cm}|}
\hline
 & Network & Accuracy & Parameters\\ \hline
 & GoogleNet {[}17{]} & 0.9370 &5M \\    \cline{2-4} 
 & VGG16 {[}18{]}     & 0.9350 & 26M \\ \cline{2-4} 
 & HOG+SVM {[}19{]}   & 0.9260           & N.A.           \\ \cline{2-4} 
\multirow{-4}{*}{Other Works } & Human {[}20{]}     & 0.8350   & N.A.                              \\ \hline
 & Sparse (ReLU)      & 0.9347 & \\ \cline{2-3} 
 & Linear      & 0.9369  & \\ \cline{2-3} 
 & Silhouette    & \textbf{0.9375} & \\ \cline{2-3} 
 & Center     & 0.9371  & \\ \cline{2-3} 
 & Contr. Center     & 0.9368  & \\ \cline{2-3} 
\multirow{-6}{*}{Our Network}  & DEFRAG &
{\textcolor{red}{\bf 0.9393}} & \multirow{-6}{*}{1.4M }\\ \hline
\end{tabular}
\end{table}

\begin{figure}[tbh]
\centering
\begin{tabular}{cc}
\includegraphics[width=45mm]{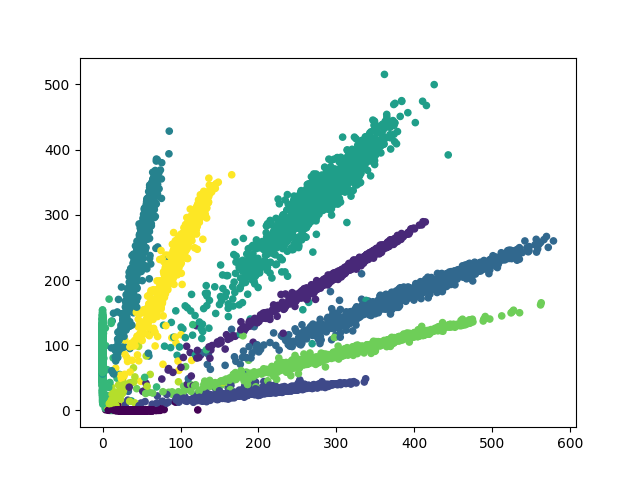}&\includegraphics[width=40mm]{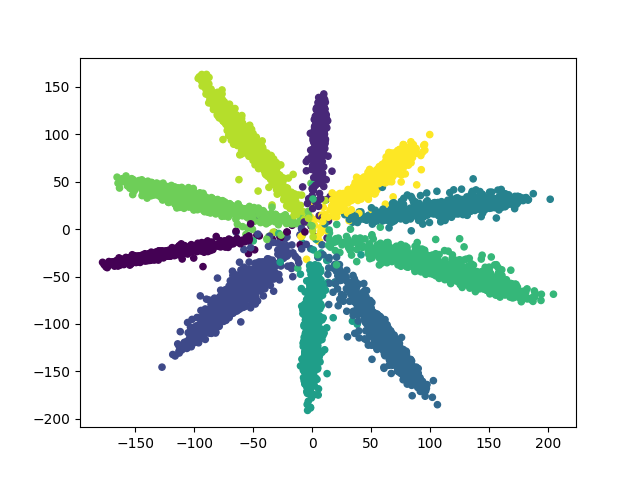}\\
(a) & (b) \\
\includegraphics[width=40mm]{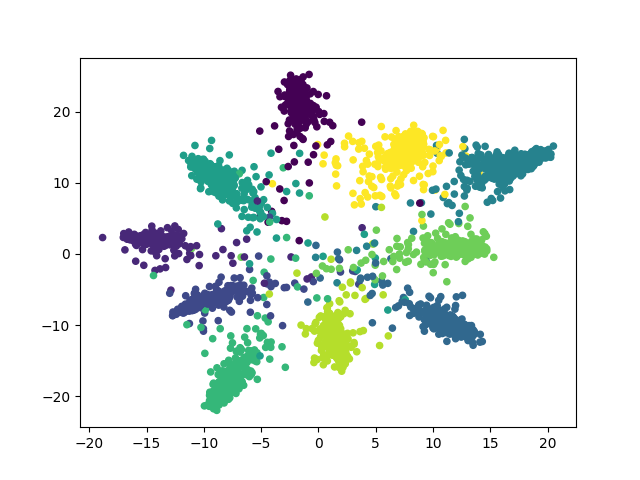}&\includegraphics[width=40mm]{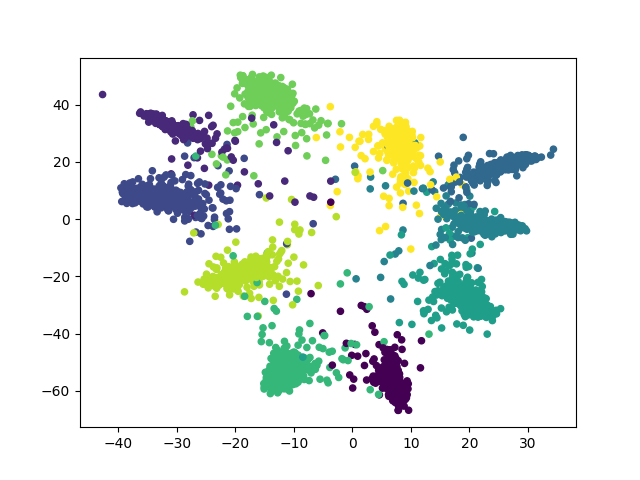}\\
(c) & (d) \\
\multicolumn{2}{c}{\includegraphics[width=50mm]{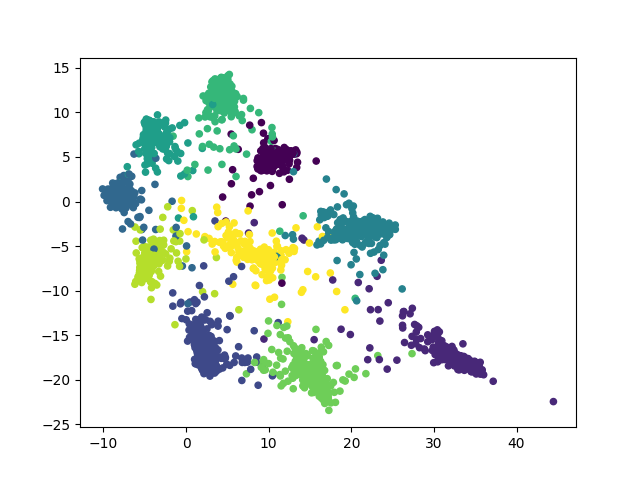}}\\
\multicolumn{2}{c}{(e)}\\
\end{tabular}
\caption{Visualization of feature representations in 2D trained on the MNIST dataset using a combination of Classification Loss and Auxiliary Loss.  Features are learned using 
(a) Softplus activation from classification loss only; (b) linear activation function; (c) center loss \cite{wzlq16}; (d) contrastive center loss \cite{qs17}; (e) our DEFRAG method.}
\label{fig:full2DLoss}
\end{figure}

\subsection{Qualitative MNIST Experiments}
\label{ssec:qualMNIST}
The MNIST dataset serves as a useful tool in understanding the behavior of the network. In this section, we qualitatively investigate the clustering behavior of the proposed method. To visualize the clustering behavior the dimensions feature representation was reduced to 2 and the resulting features were plotted in Fig. \ref{fig:full2DLoss}. The X-axis corresponds to the response of the first unit and the Y-axis corresponds to the second unit. 

A few observations can be made on these results, although is hard to draw any firm conclusions from these plots. Firstly, it is clear that the SoftPlus activation function is not ideal for classification; this was confirmed by the results where the SoftPlus implementation achieved only 88.6\% accuracy on the test set while the other methods achieved accuracies of over 99.0\%. 

Secondly, the plots of Figure \ref{fig:full2DLoss} show that auxiliary cluster losses have a significant impact on the clustering behavior of the network. The linear implementation forms linearly separable clusters for each class, however, the intra-class variance is much higher than the inter-class separation. We measured the separability of each method by both the Silhouette score or the ratio of average inner to intra class distances. In both of these metrics the DEFRAG method out performed all other methods. The DEFRAG method outperformed the center loss, the second best method, with a 16\% reduction in the Silhouette score and 13\% in the distance ratio. 

\section{Conclusions}
\label{sec:conc}
In this work, we present a new method for training deep neural networks that focuses on training better feature representations with ideal clustering behavior and can be more effectively compared using the standard $L_2$ Norm. The proposed DEFRAG method consists of a new auxiliary loss functions, Silhouette loss, and a retraction step which ensures linear independence of each dimension in the feature representation. We demonstrate performance improvements in terms of increased accuracy on multiple datasets using a smaller network, achieving state of the art on the fashion MNIST dataset.  

Our work demonstrates how by combining an auxiliary loss functions that encourages ideal clustering behavior and a orthogonalization step that ensures the feature space is $R^N$.  Clustered learned features render small networks more effective for classification applications. 

\bibliographystyle{IEEEbib}
\bibliography{bib.bib}

\end{document}